\documentclass[letterpaper, 10 pt, conference]{ieeeconf}  % Comment this line out
\IEEEoverridecommandlockouts                              % This command is only
\usepackage{graphicx} % for pdf, bitmapped graphics files
\usepackage{epsfig} % for postscript graphics files
\usepackage{mathptmx} % assumes new font selection scheme installed
\usepackage{times} % assumes new font selection scheme installed
\usepackage{amsmath} % assumes amsmath package installed
\usepackage{amssymb}  % assumes amsmath package installed
\usepackage{bm}
\usepackage{algorithm}
\usepackage{algorithmic}
\newcommand{\argmax}{\mathop{\rm arg~max}\limits}

\title{\LARGE \bf
Networked Control of Nonlinear Systems under Partial Observation Using Continuous Deep Q-Learning}

\author{Junya Ikemoto and Toshimitsu Ushio% <-this % stops a space
\thanks{This work was partially supported by JST-ERATO HASUO Project Grant Number JPMJER1603, Japan and JST-Mirai Program Grant Number JPMJMI18B4, Japan.}% <-this % stops a space
\thanks{J.\ Ikemoto and T.\ Ushio are with the Graduate School of Engineering Science, Osaka University, Toyonaka, Osaka, 560-8531, Japan
        {\tt\small ikemoto@hopf.sys.es.osaka-u.ac.jp} %%%}%
        {\tt\small ushio@sys.es.osaka-u.ac.jp}}%
}

\begin{document}

\maketitle
\thispagestyle{empty}
\pagestyle{empty}

%%%%%%%%%%%%%%%%%%%%%%%%%%%%%%%%%%%%%%%%%%%%%%%%%%%%%%%%%%%%%%%%%%%%%%%%%%%%%%%%
\begin{abstract}

In this paper, we propose a design of a model-free networked controller for a nonlinear plant whose mathematical model is unknown. In a networked control system, the controller and plant are located away from each other and exchange data over a network, which causes network delays that may fluctuate randomly due to network routing.  So, in this paper, we assume that the current network delay is not known but the maximum value of fluctuating network delays is known beforehand. Moreover, we also assume that the sensor cannot observe all state variables of the plant. Under these assumption, we apply continuous deep Q-learning to the design of the networked controller. Then, we introduce an extended state consisting of a sequence of past control inputs and outputs as inputs to the deep neural network. By simulation, it is shown that, using the extended state, the controller can learn a control policy robust to the fluctuation of the network delays under the partial observation. 
\end{abstract}

%%%%%%%%%%%%%%%%%%%%%%%%%%%%%%%%%%%%%%%%%%%%%%%%%%%%%%%%%%%%%%%%%%%%%%%%%%%%%%%%
\section{INTRODUCTION}
Reinforcement learning (RL) is one of theoretical frameworks in machine leaning (ML) and a dynamic programing-based learning approach to search of an optimal control policy \cite{Sutton}. RL is useful to design a controller for a plant whose mathematical model is unknown because RL is based on a model-free learning approach, that is, we can design the controller without plant's model. In RL, a controller that is a learner interacts with the plants and updates its control policy. The main goal of RL is to learn the control policy that maximizes the long-term rewards. RL has been applied to various control problems \cite{Lewis1}-\cite{Hiromoto}. 

Furthermore, to design a controller for a complicated system, we often use function approximations. It is known that policy gradient methods \cite{SPG,DPG} with function approximations are especially useful in control problems, where states of plants and control inputs are continuous values. Recently, deep reinforcement learning (DRL) has been actively researched. In DRL, we combine conventional RL algorithms with deep neural networks (DNNs) as high-performance function approximators \cite{DQN}-\cite{A3C}. In \cite{Masuda}, DRL is applied to a parking problem of a 4-wheeled vehicle that is a nonholonomic system. In \cite{Sangiovanni}, DRL is applied to controlling robot manipulators. Moreover, applications of DRL to networked control systems (NCSs) have also been proposed \cite{Deep_CAS}, \cite{DRL_ETC}, \cite{Ikemoto}. NCSs have been much attention to thanks to the development of network technologies. In NCSs, the controller and the plant are located away from each other. The controller computes control inputs based on outputs observed by the sensor, and sends them to the plant via a network. In \cite{Deep_CAS}, DRL is applied to control-aware scheduling of NCSs consisting of multiple subsystems operating over a shared network. In \cite{DRL_ETC}, DRL is applied to event trigger control (ETC) \cite{ETC}. However, in \cite{Deep_CAS} and \cite{DRL_ETC},  transmission delays over the network are not considered. One of the problems of NCSs is that there are network delays in the exchange of data between the controller and the plant. In the case where the network delays are constant and parameters of the network delays are known, we can design the networked controller considering the network delays. However, practically, it is difficult to identify the network delays beforehand. Moreover, the network delays may fluctuate due to the network routing. Thus, in \cite{Ikemoto}, we assume that the sensor can observe all state variables of the plant and  proposed the design of networked controller with network delays using a DRL algorithm. In general, however, the sensor cannot always observe all of them. In RL, the partial observation often degrades learning performances of the controllers.  

 In this paper, we consider the following networked control system;
\begin{itemize}
\item The plant is a nonlinear system whose mathematical model is unknown.
\item Network delays fluctuate randomly due to the network routing, where the maximum value of them is known beforehand.
\item The sensor cannot observe all state variables of the plant.
\end{itemize}
% In \cite{Fujita}, RL is applied to overcome uncertain network delays based on the idea in \cite{Lewis3}. However, in \cite{Fujita}, it is assumed that the plant is a linear system and network delays are constant. In \cite{Ikemoto}, we overcome a nonlinearity of the plant and fluctuating network delays with the continuous deep Q-learning algorithm. In this paper, as in \cite{Ikemoto}, we also apply the continuous deep Q-learning algorithm. Moreover, to overcome a partial observation, we introduce an extended state consisting of past control inputs and outputs and use it as an input of a DNN, which is also based on the key idea in \cite{Lewis3}.
Under the above assumptions, we propose a networked controller with a DNN using the continuous deep Q-learning algorithm \cite{NAF}. Then, we introduce an extended state consisting of both past control inputs and outputs of the plant as inputs to the DNN.  

The paper is organized as follows. In Section II, we review continuous deep Q-learning.  In Section III, we propose a networked controller using a DNN under the above three assumptions. In Section IV, by simulation, we apply the proposed learning algorithm to a networked controller for stabilizing a Chua circuit under the fluctuating network delays and the partial observation.  In Section V, we conclude the paper.

%%%%%%%%%%%%%%%%%%%%%%%%%%%%%%%%%%%%%%%%%%%%%%%%%%%%%%%%%%%%%%%%%%%%%%%%%%%%%%%%
\section{PRELIMINARIES}
This section reviews RL and continuous deep Q-learning that is one of DRL algorithms.
\subsection{Reinforcement Learning (RL)}
The main goal of RL is for a controller to learn its optimal control policy by trial and error while interacting with a plant. 

Let $\mathcal{X}$ and $\mathcal{U}$ be the sets of states and control inputs of the plant, respectively. The controller receives the immediate reward $r_{k}$ by the following function $R:\mathcal{X}\times\mathcal{U}\times\mathcal{X}\to\mathbb{R}$.
\begin{eqnarray}
r_{k}=R(\bm{x}_{k},\bm{u}_{k},\bm{x}_{k+1}),  
\end{eqnarray}
where $\bm{x}_k$ and $\bm{u}_k$ are the state and the control input at discrete-time $k \in \mathbb{N}$. In RL, it is necessary to evaluate the policy based on long-term rewards. Thus, the value function and Q-function are defined as follows.
\begin{eqnarray}
  V^{\mu}(\bm{x}) &=& \mathbb{E}\left[ \sum_{n=0}^{\infty}\gamma^{n}r_{n+k}|\bm{x}_{k} = \bm{x} \right],\\
Q^{\mu}(\bm{x},\bm{u}) &=& \mathbb{E}\left[ \sum_{n=0}^{\infty}\gamma^{n}r_{n+k}|\bm{x}_{k} = \bm{x},\bm{u}_{k} = \bm{u} \right],
\end{eqnarray}
where $\mu: \mathcal{X} \to \mathcal{U}$ is the evaluated control policy, that is, the input $\bm{u}$ at the state $\bm{x}$ is determined by $\bm{u}=\mu(\bm{x})$, and $\gamma\in[0,1)$ is the discount factor to prevent the divergence of the long-term rewards.

In the Q-learning algorithm, the controller indirectly learns the following greedy deterministic policy $\mu$ through updating the Q-function.  
\begin{eqnarray}
\mu(\bm{x}_{k})=\argmax_{\bm{u}\in\mathcal{U}}Q(\bm{x}_{k},\bm{u}). 
\end{eqnarray}
\subsection{Continuous Deep Q-learning with Normalized Advantage Function}
\begin{figure}
\centering
	\includegraphics[width=8.7cm]{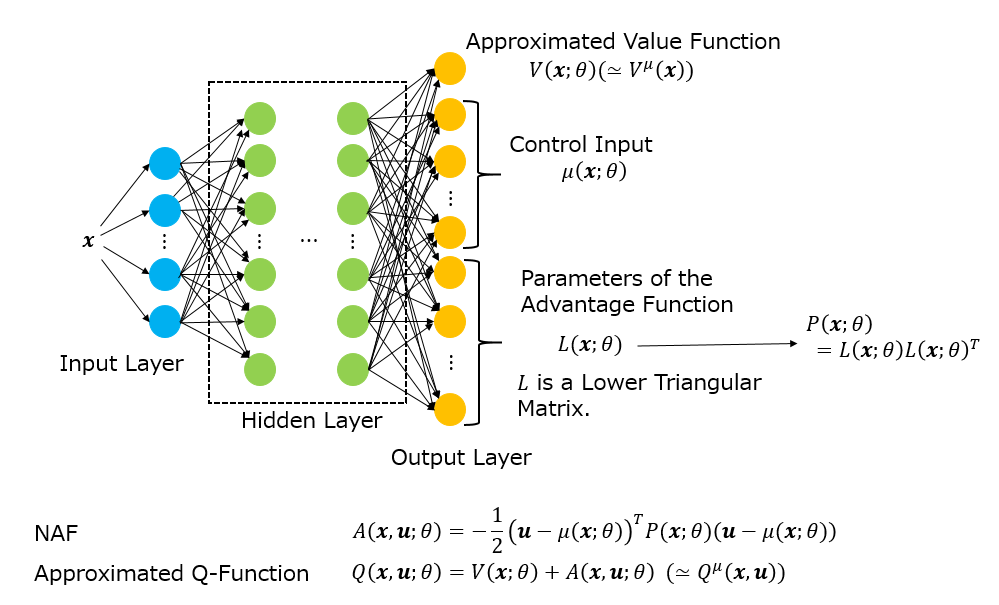}
	\caption{Illustration of a DNN for the continuous deep Q-learning with a NAF. The DNN outputs the approximated value $V(\bm{x};\theta)$, the optimal action $\mu(\bm{x};\theta)$, and the parameters of the NAF for the inputted state $\bm{x}$. }
	\label{DNN}
\end{figure}

To implement the Q-learning algorithm for plants whose state and control input are continuous, Gu \textit{et al}. proposed a parameterized quadratic function $A(\bm{x},\bm{u};\theta)$, called a normalized advantage function (NAF) \cite{NAF}, satisfying the following equations, where $\theta$ is a parameter vector of the DNN.
\begin{eqnarray}
 Q(\bm{x},\bm{u};\theta) &=& V(\bm{x};\theta)+A(\bm{x},\bm{u};\theta), \label{eqn:NAF1} \\
A(\bm{x},\bm{u};\theta) &=& -\frac{1}{2}(\bm{u}-\mu(\bm{x};\theta))^{T}P(\bm{x};\theta)(\bm{u}-\mu(\bm{x};\theta)),  \label{eqn:NAF2} \nonumber
 \\
\end{eqnarray}
where $P(\bm{x};\theta)$ is a positive definite matrix and $V(\bm{x};\theta)$ and $Q(\bm{x},\bm{u};\theta)$ are approximaitors to Eqs. (2) and (3), respectively. $\mu(\cdot;\theta)$ computes the optimal control input instead of Eq. (4), where the control input maximizes the approximated Q-function $Q(\bm{x},\bm{u};\theta)$ instead of Eq.(3). In the other words, the approximated Q-function is divided into an action-dependent term and an action-independent term, and the action-dependent term is expressed by the quadratic function with respect to the action. From Eqs. (\ref{eqn:NAF1}) and (\ref{eqn:NAF2}), when $\bm{u}=\mu(\bm{x};\theta)$, the Q-function is maximized with respect to the action $\bm{u}\in\mathcal{U}$ and we have
\begin{equation}
V(\bm{x};\theta)=\max_{\bm{u}\in\mathcal{U}}Q(\bm{x},\bm{u};\theta).
\end{equation}
\ \ We show an illustration of a DNN for continuous deep Q-learning in Fig.\ \ref{DNN}. The outputs of the DNN consist of the approximated value $V(\bm{x};\theta)$, the optimal control input $\mu(\bm{x};\theta)$, and the parameters that constitute the lower triangular matrix $L(\bm{x};\theta)$, where the diagonal terms are exponentiated. Moreover, the positive definite matrix $P(\bm{x};\theta)$ is given by $L(\bm{x};\theta)L(\bm{x};\theta)^{T}$.\\

%%%%%%%%%%%%%%%%%%%%%%%%%%%%%%%%%%%%%%%%%%%%%%%%%%%%%%%%%%%%%%%%%%%%%%%%%%%%%%%%
\section{CONTINUOUS DEEP Q-LEARNING-BASED NETWORK CONTROL}
\subsection{Networked Control System}
	\begin{figure}
      \centering
      \includegraphics[width=8.7cm]{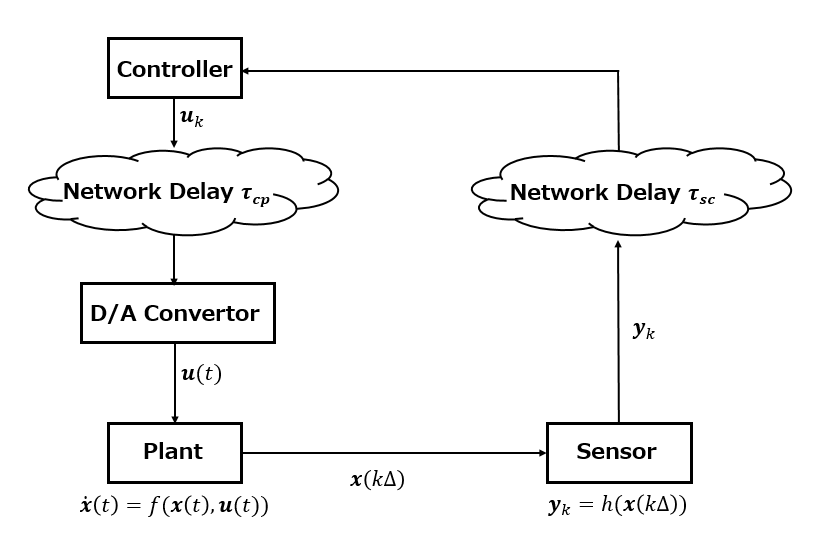}
      \caption{Block diagram of NCSs problem in which the network delays are caused by transmissions of control inputs and observed outputs.}
      \label{Diagram}
   \end{figure}
We consider the networked control of the following nonlinear plant as shown in Fig.\ \ref{Diagram}.
\begin{eqnarray}
\dot{\bm{x}}(t) &=&f(\bm{x}(t),\bm{u}(t)), \\
\bm{y}_{k} &=& h(\bm{x}(k\Delta)),
\end{eqnarray}
where
\begin{itemize}
\item $\bm{x}(t)\in\mathcal{X}\subseteq{\mathbb{R}^{n}}$ is the state of the plant at time $t\in\mathbb{R}$,
\item $\bm{u}(t)\in\mathcal{U}\subseteq\mathbb{R}^{m}$ is the control input of the plant at time $t\in\mathbb{R}$ and the $k$-th updated control input computed by the digital controller is denoted by $\bm{u}_{k}$,
\item $\bm{y}_{k}\in\mathcal{Y}\subseteq{\mathbb{R}^{p}}\ (p<n)$ is the $k$-th output of the plant observed by the sensor,
\item $\Delta$ is the sampling period of the sensor,  
\item $f:\mathbb{R}^{n}\times\mathbb{R}^{m}\to\mathbb{R}^{n}$ describes the mathematical model of the plant, but it is assumed to be unknown, and
\item $h:\mathbb{R}^{n}\to\mathbb{R}^{p}$ is the output function of the plant that is characterized by the sensor.
\end{itemize}
The discrete-time control input $\bm{u}_k$ is sent to the D/A converter and held until the next input is received. The plant and the digital controller are connected by information networks and there are two types of network delays: One is caused by transmissions of the observed outputs from the sensor to the controller while the other is by those of the updated control inputs from the digital controller to the plant. The former and the latter delay at discrete-time $k$ are denoted by $\tau_{sc,k}$ and $\tau_{cp,k}$, respectively. Then, for $k\Delta+\tau_{sc,k}+\tau_{cp,k}\le t < (k+1)\Delta+\tau_{sc,k+1}+\tau_{cp,k+1}$,
\begin{eqnarray}
\bm{u}(t)=\bm{u}_{k}. 
\end{eqnarray}
We assume that the packet loss does not occur in the networks and all data are received in the same order as their sending order. We assume that both network delays are upper bounded and the maximum values of them are known beforehand.
\subsection{Extended State}
We assume that the maximum values of $\tau_{sc,k}$ and $\tau_{cp,k}$ are known and let $\max (\tau_{sc,k})=a\Delta$ and $\max(\tau_{cp,k})=b\Delta$ ($a,b\in\mathbb{N}$ and $a+b=\tau$). 

\begin{figure*}
\centering
	\includegraphics[width=16.5cm]{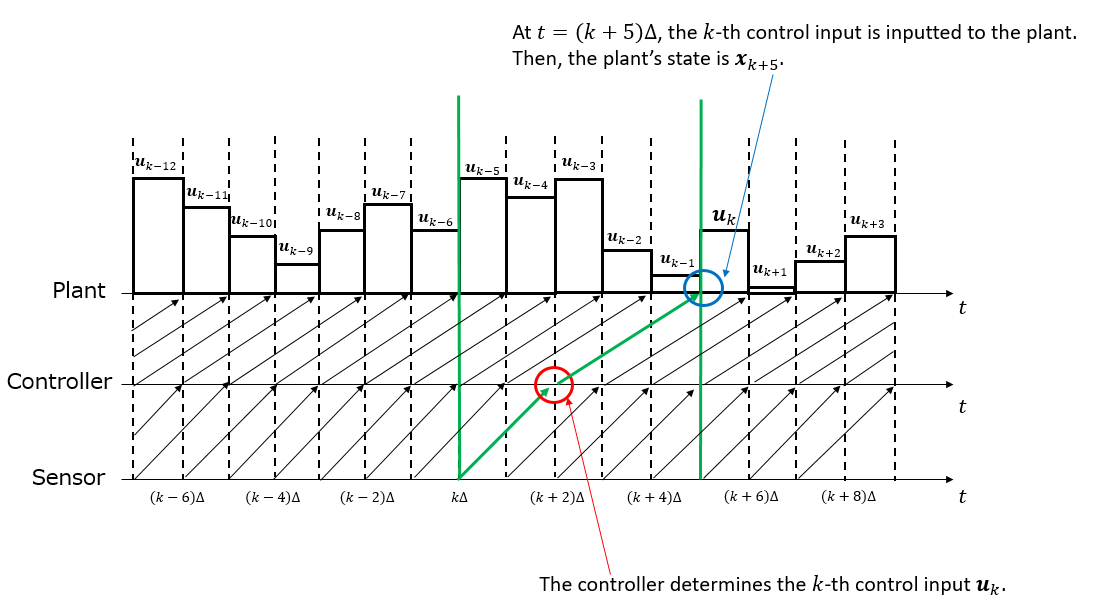}
	\caption{We consider the case where $\max(\tau_{sc,k})=2\Delta$ and $\max(\tau_{cp,k})=3\Delta$. If the all network delays are maximum, the controller needs the past control input sequence $\{\bm{u}_{k-1},\ \bm{u}_{k-2},\ ...,\ \bm{u}_{k-5}\}$ to estimate $\bm{x}_{k+5}$ and compute the $k$-th control input $\bm{u}_{k}$.}
	\label{Delay}
\end{figure*}

First, we consider randomly fluctuated delays. The sensor observes the $k$-th output $\bm{y}_{k}$ at $t=k\Delta$. The controller receives the $k$-th output $\bm{y}_{k}$ and computes the $k$-th control input $\bm{u}_{k}$ at $t=k\Delta+\tau_{sc,k}$. The $k$-th control input $\bm{u}_{k}$ is inputted to the plant at $t=k\Delta+\tau_{sc,k}+\tau_{cp,k}$. Then, the controller must estimate the state $\bm{x}(k\Delta+\tau_{sc,k}+\tau_{cp,k})$ and compute $\bm{u}_{k}$. Since $\tau_{sc,k}+\tau_{cp,k} \leq \tau\Delta$, we use the last $\tau$ control inputs that the controller needs in the worst case to estimate the state of the plant as shown in Fig. \ref{Delay}. Thus, in [20], we introduced the extended state $\bm{z}_{k}=[\bm{x}_{k},\ \bm{u}_{k-1},\ \bm{u}_{k-2},\ ...,\ \bm{u}_{k-\tau}]^{T}$ and use it as the input of a DNN. We also use the past control input sequence. However, in this paper, the sensor cannot observe all state variables $\bm{x}_{k}$. 

\begin{figure}
\centering
	\includegraphics[width=8.5cm]{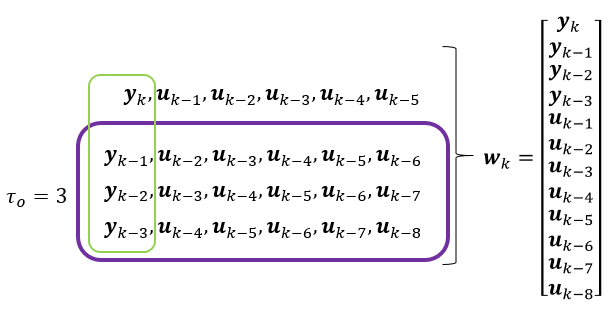}
	\caption{In the case where the sensor cannot observe all state variables of the plant, we use past control inputs and outputs as the extended state. For example, we set $\tau_{o}=3$. Then, we use the past control inputs $\{\bm{u}_{k-1},\bm{u}_{k-2},...,\bm{u}_{k-8}\}$ and outputs $\{\bm{y}_{k-1},\bm{y}_{k-2},\bm{y}_{k-3}\}$ as the extended state $\bm{w}_{k}$.}
	\label{Partial}
\end{figure}

Second, we consider a partial observation. In \cite{Data-Based}, Aangenent \textit{et al.} proposed a data-based optimal control method using past control input and output sequence in the case where the plant is a linear, controllable, and observable system. Then, the length of the sequence must be larger than the observability index $K$ of the plant, where $K\le n \le Kp$. Similarly, in this paper, we also use the past control input and output sequence $\{\bm{y}_{k-1},\ \bm{y}_{k-2},\ ...,\ \bm{y}_{k-\tau_{o}},\ \bm{u}_{k-(\tau+1)},\ \bm{u}_{k-(\tau-2)},\ ...,\ \bm{u}_{k-(\tau+\tau_{o})}\}$ to estimate the state $\bm{x}_{k}$ as shown in Fig.\ \ref{Partial}, where the hyper parameter $\tau_{o}\in\mathbb{N}$ is selected beforehand. Although there is no theoretical guarantee, we select $\tau_{o}$ conservatively such that $\tau_o \geq n$
.  We define the following extended state $\bm{w}_{k}$.

\begin{figure}
\centering
	\includegraphics[width=8.5cm]{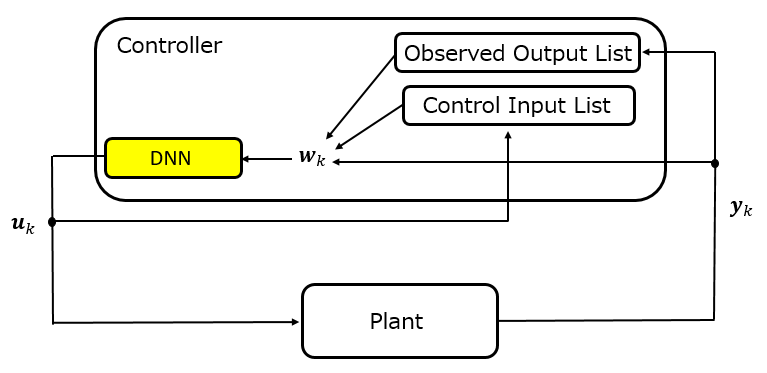}
	\caption{The network controller is designed with a DNN. When the controller receives the $k$-th observed output $\bm{y}_{k}$ at $t=k\Delta+\tau_{sc,k}$, it computes the $k$-th control input $\bm{u}_{k}$ based on the extended state $\bm{w}_{k}$. Then, the input of the DNN is $\bm{w}_{k}$. The parameter vector of the deep neural network for the controller is optimized by the continuous deep Q-leaning algorithm \cite{NAF}. }
	\label{Controller}
\end{figure}

\begin{eqnarray}
\bm{w}_{k}=\left[
	\begin{array}{ccccccc}
	\bm{y}_{k}\\
	\vdots\\
	\bm{y}_{k-\tau_{o}}\\
	\bm{u}_{k-1}\\
	\bm{u}_{k-2}\\
	\vdots\\
	\bm{u}_{k-(\tau+\tau_{o})}
\end{array}
\right]\in \mathbb{R}^{p\tau_{o}+m(\tau+\tau_{o})}.
\end{eqnarray}
Thus, we design the networked controller with a DNN as shown in Fig.\ \ref{Controller}.  
\begin{algorithm}[h]               
\caption{Continuous Deep Q-learning with the NAF and the $\tau$-extended state of the networked control systems}    
\label{alg2}                          
\begin{algorithmic}[1]
\STATE Select the length of the past output sequence $\tau_0$
\STATE Initialize the replay memory $D$.
\STATE Randomly initialize the main Q network with weights $\theta$.
\STATE Initialize the target network with weights $\theta^{-} = \theta$.
\FOR{episode$=1,...,M$}
\STATE Initialize the initial state $\bm{x}_{0}\sim p(\bm{x}_{0})$.
\STATE Receive the initial observed output $\bm{y}_{0}$.
\STATE Memorize the observed output $\bm{y}_{0}$.
\STATE Generate the initial extended state $\bm{w}_{0}$, where $\bm{u}_{i} = 0 (i<0)$.
\STATE Initialize a random process $\mathcal{N}$ for action exploration.
\FOR{$k=0,...,K$}
\IF{$k>0$}
\STATE Receive the $k$-th output $\bm{y}_{k}$.
\STATE Memorize the observed output $\bm{y}_{k}$.
\STATE Generate the extended state $\bm{w}_{k}$ with past control inputs and outputs.
\STATE Return the reward $r_{k-1}=R(\bm{w}_{k-1},\bm{u}_{k-1},\bm{w}_{k})$.
\STATE Store the transition $(\bm{w}_{k-1},\bm{u}_{k-1},\bm{w}_{k},r_{k-1})$ in $D$.
\ENDIF
\STATE Determine the control input $\bm{u}_{k} = \mu(\bm{w}_{k};\theta) + \mathcal{N}_{k}$ and send the  control input to the plant.
\STATE Memorize the control input $\bm{u}_{k}$.
\IF{$k\%k_p=0$}
\FOR{iteration$=1,...,I$}
\STATE Sample a random minibatch of $N$ transitions $(\bm{w}^{(n)},\bm{u}^{(n)},\bm{w}'^{(n)},r^{(n)}),\ n=1,...,N$ from $D$.
\STATE Set $t^{(n)}=r^{(n)}+\gamma V(\bm{w}'^{(n)};\theta^{-})$.
\STATE Update $\theta$ by minimizing the loss: \\
\ \ $J(\theta)=\frac{1}{N}\sum_{n=1}^{N} (t^{(n)}-Q(\bm{w}^{(n)},\bm{u}^{(n)};\theta))^{2}$.
\STATE Update the target network:\\
\ \ $\theta^{-} \gets \beta\theta + (1-\beta)\theta^{-}$.
\ENDFOR
\ENDIF
\ENDFOR
\ENDFOR
\end{algorithmic}
\end{algorithm}
\subsection{DRL algorithm}
The parameter vector of the DNN for the controller is optimized by the continuous deep Q-leaning algorithm \cite{NAF}. The input to the DNN is the extended state $\bm{w}_{k}$. Shown in Algorithm 1 is the proposed learning algorithm. In the same way as the DQN algorithm \cite{DQN}, we use the experience replay and the target network. The parameter vectors of the main network and the target network are denoted by $\theta$ and $\theta^{-}$, respectively. For the update of $\theta$, the following TD error is used.
\begin{eqnarray}
J(\theta)&=&\left( t_{k} - Q(\bm{w}_{k},\bm{u}_{k};\theta) \right)^{2}, \\
t_{k} &=& r_{k} + \gamma V(\bm{w}_{k};\theta^{-}).
\end{eqnarray}
Moreover, for the update of $\theta^{-}$, the following soft update is used.
\begin{equation}
\theta^{-} \gets \beta\theta + (1-\beta)\theta^{-},
\end{equation}
where $\beta$ is given as a very small positive real number.

The exploration noises $\mathcal{N}_{k}\ (k=1,2,...)$ are generated under a given random process $\mathcal{N}$.

%%%%%%%%%%%%%%%%%%%%%%%%%%%%%%%%%%%%%%%%%%%%%%%%%%%%%%%%%%%%%%%%%%%%%%%%%%%%%%%%
\section{SIMULATION}
We apply the proposed controller to a stabilization of a Chua circuit as follows. We set the sampling period to $\Delta=2^{-4}(s)$. Moreover, we assume that the terminal of a leaning episode is at $t=12.0 (s)$. 
\subsection{Chua circuit}
\begin{figure}
	\centering
	\includegraphics[width=8.0cm]{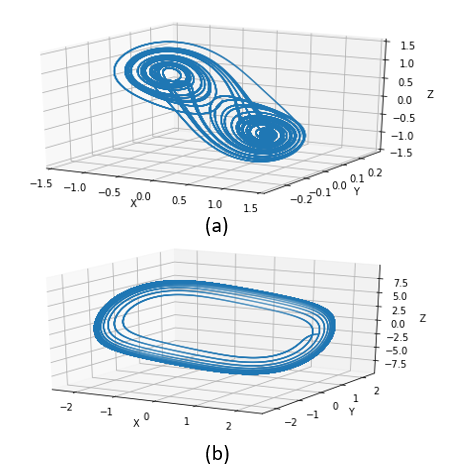}
	\caption{The trajectory of a Chua circuit is depending on an initial state. (a) The initial state is $[-0.2, 0.1, -0.1]$ and its behavior converges to a chaotic attractor. (b) The initial state is $[2.0, -1.0, 1.0]$ and its behavior converges to a limit cycle.}
	\label{Chua}
\end{figure}

The dynamics of a Chua circuit is given by
\begin{eqnarray}
\frac{d}{dt}
\left[
    \begin{array}{cccc}
      x(t)\\
	y(t)\\
	z(t)
    \end{array}
  \right]  &=&\left[
    \begin{array}{cccc}
      p_{1}(y(t)-\phi(x(t)))\\
	x(t)-y(t)+z(t)+u(t)\\
	-p_{2}y(t)
    \end{array}
  \right], 
\end{eqnarray}
where $\phi(x)=(2x^3-x)/7$. In this simulation, we assume that $p_{1}=10$ and $p_{2}=100/7$, where these parameters are unknown. The Chua circuit has a chaotic attractor and a limit cycle as shown in Fig.\ \ref{Chua}. We assume that the states $x$ and $y$ are sensed as follows.
\begin{eqnarray}
\bm{y}_{k} = \left[
    \begin{array}{cccc}
      1&0&0\\
	0&1&0
    \end{array}
  \right]
\left[
    \begin{array}{cccc}
      x(k\Delta)\\
	y(k\Delta)\\
	z(k\Delta)
    \end{array}
  \right]. 
\end{eqnarray}

We assume that equilibrium points of the Chua circuit is unknown because the controller does not know the parameters $p_{1}$ and $p_{2}$. Thus, we define the reward function in this simulation as follows. 

First, we define the reward $r_{k}^{(1)}$ based on outputs $\bm{y}_{k},\ \bm{y}_{k+1}$ and the control input $u_k$.
\begin{equation}
r_{k}^{(1)} = -(\bm{y}_{k+1}-\bm{y}_{k})^{T}\left[
    \begin{array}{cccc}
      0.8 & 0\\
	0 & 0.8
    \end{array}
  \right](\bm{y}_{k+1}-\bm{y}_{k})-1.0u_{k}^{2}.
\end{equation}

Second, we define the reward based on the past output sequence used for the extended state. 
\begin{equation}
r_{k}^{(2)} = -\sum_{i=1}^{\tau_{o}}\left\{ (\bm{y}_{k+1-i}-\bm{y}_{k-i})^{T}\left[
    \begin{array}{cccc}
      0.8 & 0\\
	0 & 0.8
    \end{array}
  \right](\bm{y}_{k+1-i}-\bm{y}_{k-i}) \right\}.
\end{equation}

Third, we define the reward based on the past control input sequence used for the extended state.
\begin{equation}
r_{k}^{(3)} = -0.15\sum_{i=1}^{\tau+\tau_{o}}\left( u_{k+1-i}-u_{k-i} \right)^2.
\end{equation}

Finally, we define the immediate reward at discrete-time $k$ as follows.
\begin{equation}
r_{k} = r_{k}^{(1)}+r_{k}^{(2)}+r_{k}^{(3)}.
\end{equation}
Thus, the goal of DRL is the stabilization of the circuit at one of equilibrium points.
\subsection{Design of the controller}
We use a DNN with four hidden layers, where all hidden layers have 128 units and all layers are fully connected layers. The activation functions are ReLU except for the output layer. Regarding the activation functions of the output layer, we use a linear function for both the $V$ unit and units for parameters of the advantage function, while we use a weighted hyperbolic tangent function for the $\mu$ unit. The size of the replay memory is $1.0\times10^{6}$ and the minibatch size is 128. The parameters of the DNN are updated 10 times per 4 discrete time steps ($I=10,\ k_p=4$) by ADAM \cite{ADAM}, where its learning stepsize is $1.25\times10^{-5}$. The soft update rate $\beta$ for the target network is 0.001, and the discount rate $\gamma$ for the Q-value is 0.99. 

For the exploration noise process, we use an Ornstein-Uhlenbeck process \cite{OUnoise}. The exploration noises are multiplied by 3.5 during the 1st to the 1000th episode, and the noise is gradually reduced after the 1001st episode. The initial state is randomly selected for each episode, where $-4.5\le x(0) \le 4.5$, $-4.5\le y(0) \le 4.5$, and $-4.5\le z(0) \le 4.5$.  
\subsection{Result}
\begin{figure}
	\centering
	\includegraphics[width=6.8cm]{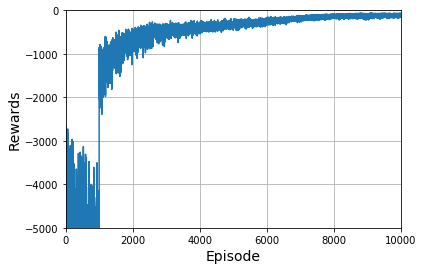}
	\caption{Learning curve. For all $k$, $\Delta\le\tau_{sc,k}\le3\Delta$ and $\Delta\le\tau_{cp,k}\le3\Delta$. The values of the vertical axis are given by the sum of $r_{k}$ between the 50th sampling and the episode's terminal for each episode.}
	\label{Result1}
\end{figure}

\begin{figure}
	\centering
	\includegraphics[width=6.8cm]{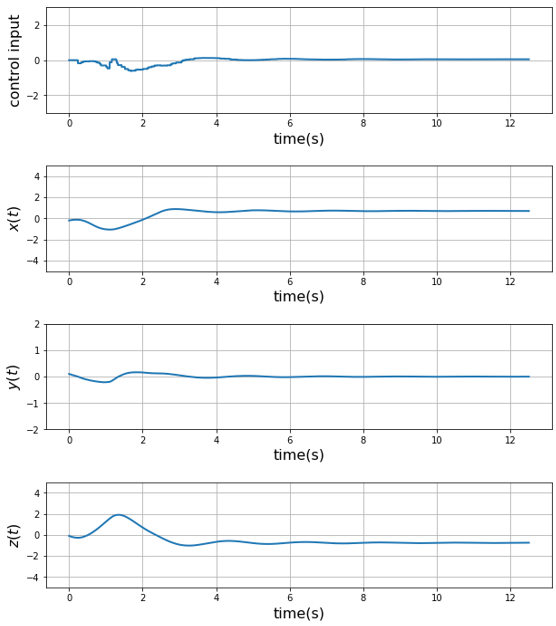}
	\caption{Time response of the Chua circuit using a control policy after 8500 episodes. For all $k$, $\Delta\le\tau_{sc,k}\le3\Delta$ and $\Delta\le\tau_{cp,k}\le3\Delta$. The parameter $\tau$ is set to 8 and the parameter $\tau_{o}$ is set 4. The initial state is $[-0.2, 0.1, -0.1]$, where its converged behavior is shown in Fig.\ 6(a).}
	\label{Result3}
\end{figure}

\begin{figure}
	\centering
	\includegraphics[width=6.8cm]{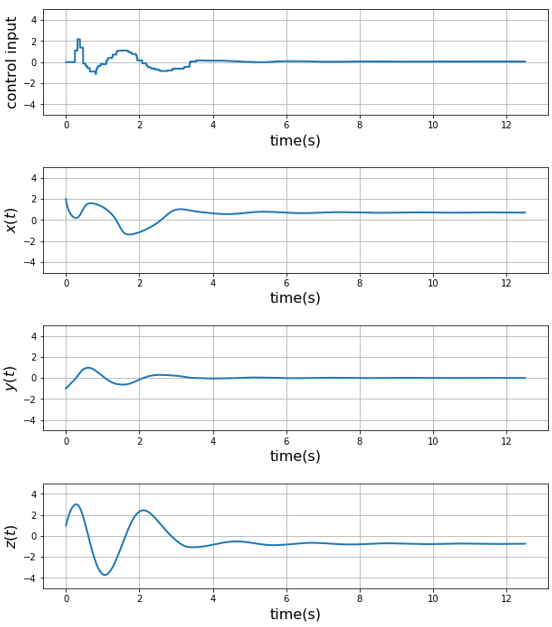}
	\caption{Time response of the Chua circuit using a control policy after 8500 episodes. For all $k$, $\Delta\le\tau_{sc,k}\le3\Delta$ and $\Delta\le\tau_{cp,k}\le3\Delta$. The parameter $\tau$ is set to 8 and the parameter $\tau_{o}$ is set to 4. The initial state is $[2.0, -1.0, 1.0]$, where its converged behavior is shown in Fig.\ 6(b).}
	\label{Result4}
\end{figure}

First, we assume that, for all $k\in\mathbb{N}$, the network delays are set to $\Delta\le\tau_{sc,k}\le 3\Delta$ and $\Delta \le \tau_{cp,k} \le 3\Delta$. These ranges are unknown. However, we assume that we know $\max(\tau_{sc,k})=4\Delta$ and $\max(\tau_{cp,k})=4\Delta$ beforehand. Thus, we set $\tau=8$. Moreover, we select $\tau_{o}=4$, that is, $\tau_{o}$ is larger than the dimension of the plant's state ($n=3$). The learning curve is shown in Fig.\ \ref{Result1}, where the values of the vertical axis, called rewards, are given by the sum of $r_{k}$ between the 50th sampling and the episode's terminal for each episode. It is shown that the controller can learn a control policy that achieves a high reward. Moreover, shown in Figs.\ \ref{Result3} and \ref{Result4} are the time responses of the Chua circuit using the control policy after 8500 episodes. It is shown that the controller that sufficiently learned the control policy using the proposed method can stabilize the circuit.

\section{CONCLUSION}
In this paper, we proposed a model-free networked controller for a nonlinear plant with network delays using continuous deep Q-learning. Moreover, the sensor cannot observe all state variables of the plant. Thus, we introduce an extended state consisting of a sequence of the past control inputs and outputs and use it as an input to a DNN. We showed the usefulness of the proposed controller by stabilizing a Chua circuit. It is future work to extend the proposed controller under the existence of packet loss and sensing noises. 
%%%%%%%%%%%%%%%%%%%%%%%%%%%%%%%%%%%%%%%%%%%%%%%%%%%%%%%%%%%%%%%%%%%%%%%%%%%%%%%%

\end{document}